\documentclass[11pt]{article}

\makeatletter
\renewcommand\@biblabel[1]{}
\makeatother
\usepackage{natbib}
\usepackage{graphicx}
\usepackage{linguex}
\usepackage{xurl}
\usepackage{amsmath}
\usepackage{appendix}
\usepackage{booktabs}
\usepackage[dvipsnames]{xcolor}
\usepackage{soul}
\usepackage[all]{nowidow}
\usepackage[a-1b]{pdfx}

\usepackage{times}

\newenvironment{sciabstract}{%
\begin{quote} \bf}
{\end{quote}}

\title{Not-So-Strange Love: Language Models and Generative Linguistic Theories are More Compatible than They Appear}

\author
{R.\ Thomas McCoy\\
\\
\normalsize{\textit{Open Peer Commentary on ``How Linguistics Learned to Stop Worrying and}} \\
\normalsize{\textit{Love the Language Models'' by Richard Futrell and Kyle Mahowald}} \\
}

\date{}

\begin{document} 

\setcitestyle{round}


\baselineskip16pt


\maketitle


%TC:ignore
\begin{sciabstract}
\cite{futrell2025linguistics} frame the success of neural language models (LMs) as supporting gradient, usage-based linguistic theories. I argue that LMs can also instantiate theories based on formal structures---the types of theories seen in the generative tradition. This argument expands the space of theories that can be tested with LMs, potentially enabling reconciliations between usage-based and generative accounts.
\end{sciabstract}
%TC:endignore

% In setting up this template for *Science* papers, we've used both
% the \section* command and the \paragraph* command for topical
% divisions.  Which you use will of course depend on the type of paper
% you're writing.  Review Articles tend to have displayed headings, for
% which \section* is more appropriate; Research Articles, when they have
% formal topical divisions at all, tend to signal them with bold text
% that runs into the paragraph, for which \paragraph* is the right
% choice.  Either way, use the asterisk (*) modifier, as shown, to

\noindent
What role does formal structure play in language acquisition? This question is at the heart of longstanding debates in linguistics. Generative linguists argue that structure guides learning, through innate predispositions for certain formal properties \citep{chomsky1993minimalism}. Usage-based linguists instead argue that the central factor is statistics of language use, with formal structure emerging instead of being innate \citep{bybee2001frequency}. \citeauthor{futrell2025linguistics} align language models (LMs) with the usage-based tradition, framing them as ``a proof of concept for gradient, usage-based theories of language.'' Here, I argue that while LMs can serve this purpose, they can also instantiate theories based on formal structures. Therefore, they are compatible with generative theories as well as usage-based ones, and their success does not uniquely vindicate one or the other. 

As \citeauthor{futrell2025linguistics} discuss, standard LMs possess many properties that align with usage-based theories, such as their statistical nature and lack of built-in structural constraints. Thus, LMs indeed provide partial support for usage-based ideas. However, this support is only partial due to an important limitation of standard LMs: as \citeauthor{futrell2025linguistics} note, LMs are typically trained on far more linguistic input than children receive. When they instead receive more realistic amounts of data, their linguistic abilities are less impressive \citep{yedetore2023poor}. Such limitations could potentially be overcome by predisposing language models toward formal linguistic structures, since a major motivation for centering formal structure is to explain how children acquire language from so little data. 

In fact, there is an approach that can distill a formal-structure-based linguistic theory into a neural network, enabling LMs to serve as empirical testing grounds for generative theories as well as usage-based ones \citep{mccoy2025modeling}. This approach uses a technique called meta-learning, in which the network is shown many languages sampled from a space of possible languages. Through this process, the network learns which sorts of languages are likely and unlikely, endowing it with a version of Universal Grammar. This information provides helpful guidance for language learning, enabling the network to learn new languages more readily. 

For example, in \cite{mccoy2020universal}, we used this approach to distill an Optimality Theory account of syllable structure \citep{princesmolensky1993ot} into a neural network, giving it learning biases that instantiated this generative theory. The resulting network could learn syllable structure patterns from just 100 examples, compared to the 20,000 examples needed by standard networks. Further, the network generalized well to novel types of examples, such as words longer than any it had seen. Thus, after having a generative theory distilled into it, the network displayed precisely the strengths that structure-centric learning theories are intended to capture: rapid learning and effective extrapolation---areas where standard neural networks struggle. In addition to this phonological case study, we have applied the same technique to syntax, showing that learning biases for certain formal language mechanisms can be distilled into LMs \citep{mccoy2025modeling}. These experiments show that we can create neural networks with a decidedly generative flavor, in that predispositions for specific formal structures drive their learning. 

The fact that LMs can instantiate generative theories raises the possibility that existing LMs might already implicitly realize aspects of these theories, further complicating the question of whether existing LM successes are driven by usage-based or structural properties. Structural biases could arise from architectural factors \citep{smolensky2022neurocompositional} or from highly structured training data such as computer code \citep{kim2024code}, which might contribute formal structural biases through a process akin to meta-learning. Whether existing LMs indeed have formal structural influences, and whether the relevant structures align with existing generative theories, are important questions for future research.

Because LMs can instantiate usage-based theories or generative ones, they create opportunities to evaluate both types of theories. Further, the gradient nature of LMs means that they are not restricted to incorporating only one type of theory but can instead combine usage-based and generative properties. Therefore, they create a pathway for exploring how we can combine the complementary strengths of these two paradigms \citep{boleda2025llms}.

As one example of synthesizing the two schools of thought, consider the aforementioned syllable structure experiment, which instantiated a generative theory inside a neural network. The instantiation's neural nature meant that distinctions that were absolute in the generative theory became graded in the neural instantiation. For instance, under the generative theory in question, certain types of languages are impossible, but our neural instantiation of the theory could still learn such languages; ``impossible'' languages required far more training data than ``possible'' ones, but they could eventually be learned. Thus, the notion of impossibility in the generative theory became improbability inside its neural instantiation. There are at least two ways that this inconsistency can be viewed as combining generative ideas and gradient, usage-based ones. First, the network can be viewed as combining the causal influence of formal structure from the generative tradition with the gradience typically associated with usage-based theories. Under this framing, the network is neither fully generative nor fully aligned with the usage-based tradition but rather combines elements of both. Alternatively, the network could be viewed as affording two levels of analysis in the sense of \cite{marr1982vision}. On one level---the level defined by the neural network, corresponding to \citeauthor{marr1982vision}'s algorithmic level---the system is gradient, while at a more abstract, idealized level---the level defined by the generative theory that the network was optimized to realize, corresponding to Marr's computational level---the system is discrete. This framing illustrates how a single system can be understood in different ways at different levels \citep{smolenskylegendre2006harmonic}, as another way in which seemingly incompatible theoretical traditions can be reconciled.

There is precedent for generative linguistics being deeply shaped by neural networks. Optimality Theory, an influential generative formalism, was motivated by neural networks, demonstrating that neural networks and generative theories can interact productively. Recent progress in neural language models creates further opportunities for advancing generative linguistics and for exploring syntheses between generative and usage-based theories.

\section*{Acknowledgments}

I am grateful to Robert Frank and Kate McCurdy for helpful discussion. Any errors are my own.

\bibliography{scibib}

\bibliographystyle{apalike}

\end{document}